# Machine Learning, Linear and Bayesian Models for Logistic Regression in Failure Detection Problems

B. Pavlyshenko

SoftServe, Inc., Ivan Franko National University of Lviv,
Lviv, Ukraine
e-mail: b.pavlyshenko@gmail.com

In this work, we study the use of logistic regression in manufacturing failures detection. As a data set for the analysis, we used the data from Kaggle competition "Bosch Production Line Performance". We considered the use of machine learning, linear and Bayesian models. For machine learning approach, we analyzed XGBoost tree based classifier to obtain high scored classification. Using the generalized linear model for logistic regression makes it possible to analyze the influence of the factors under study. The Bayesian approach for logistic regression gives the statistical distribution for the parameters of the model. It can be useful in the probabilistic analysis, e.g. risk assessment.

## INTRODUCTION

It is important to monitor closely the manufacturing parts, as they progress through the manufacturing processes. Bosch records data at every step along its assembly lines. They have the ability to apply advanced analytics to improve these manufacturing processes. However, the intricacies of the data and complexities of the production line pose some problems for current methods. In the Kaggle competition "Bosch Production Line Performance"[1], Bosch is challenging Kagglers to predict internal failures using thousands of measurements and tests made for each component along the assembly line. This would enable Bosch to bring quality products at lower costs to the end user. The peculiarity of this competition is that the classification classes are highly imbalanced. The task of internal failures prediction can be considered as a kind of logistic regression problem. Classifications are evaluated using Matthews correlation coefficient (MCC) between the predicted and the observed response. The MCC is given by

$$MCC = \frac{(TP \cdot TN) - FP \cdot FN}{\sqrt{(TP+FP)(TP+FN)(TN+FP)(TN+FN)}}$$

where $TP$ is the number of true positives, $TN$ is the number of true negatives, $FP$ is the number of false positives, and $FN$ is the number of false negatives. The data for this competition represents the measurements of parts, as they move through Bosch's production lines. Each part has a unique Id. The goal is to predict which parts will fail quality control (represented by a Response=1). The data set contains an extremely large number of anonymized features. The goal of this study is to consider different approaches for time manufacturing parts of internal failure modeling. Machine learning algorithms make it possible to find patterns in the data. This approach can give high precision in the logistic regression. We can define the importance of each feature but it cannot give us any information about the influence of each factor. The combination of parametric and non-parametric models can give us more complete analysis with complimentary approaches. As parametric models, we can consider the generalized linear model for logistic regression. Given the features in the analyzed data sets are anonymized, we can consider the internal failures prediction as a stochastic process. If we have sales distribution, we can calculate the value at risk (VaR), which is one of risk assessment features. To find distributions of model parameters Bayesian inference approach can be used.

## MACHINE LEARNING MODEL

For machine learning approach, we used a gradient boosting classification method implemented in the XGBoost classifier. To perform the analysis, we used R package "xgboost" (short term for eXtreme Gradient Boosting) which is an efficient and scalable implementation of gradient boosting framework [2,3,4]. The package includes efficient linear model solver and a tree-learning algorithm. Our first step was to prepare our data. We merged data frames with numeric, categorical, and date features. We also constructed a new feature denoting the time at which the part was on the manufacturing line. We calculated this feature as the difference between minimum and maximum time in the date features sets. Taking into account the large amount of the data set, we used the undersampling approach for logistic regression with highly imbalanced classes. We retain the samples with positive response 1 without changes but the number of samples with negative response 0 was reduced essentially using random sampling. For categorical features, one-hot encoding was used. To the data set, obtained in such a way, we applied the XGBoost classifier. The total number of features was more than 4000. We defined the most important features by applying XGBoost classifier to the small subset of the whole training set. For our next investigation, we took 500 top most important features. For the validation data set, we took 25% of training data set samples. The results of the classification were the probability for positive responses. Fig. 1 shows the most important features and their gain values. Fig.2 shows the ROC curve for the classification results. Calculated AUC is 0.753. To get the binary values of response, we need to apply a threshold for received probabilities. If the probability is larger than the threshold, then the response takes on the value 1, otherwise the value is 0. Fig. 3 shows Matthews correlation coefficient for the logistic regression for different values of the threshold. Fig.4 shows the Matthews correlation coefficient for the subsets of samples received by undersampling. The results have slight differences for different samples subsets which can be taken into account on the second level, using probabilistic Bayesian model.

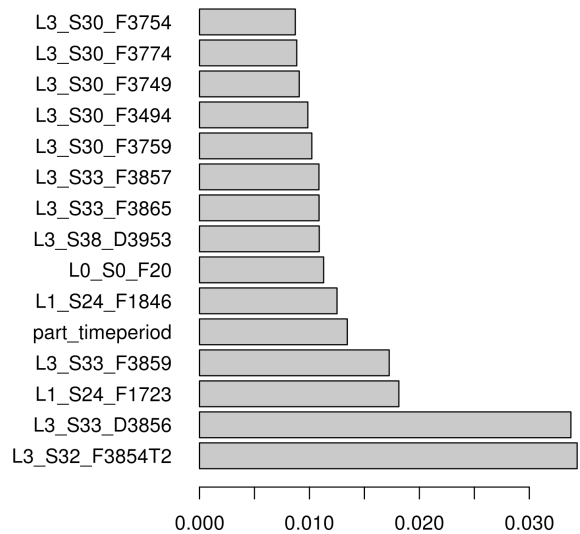

Fig.1. The most important features and their gain values.

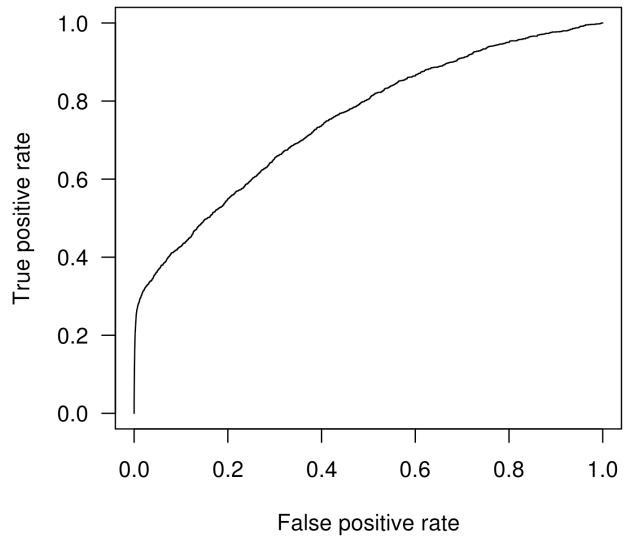

Fig.2. ROC curve for classification results

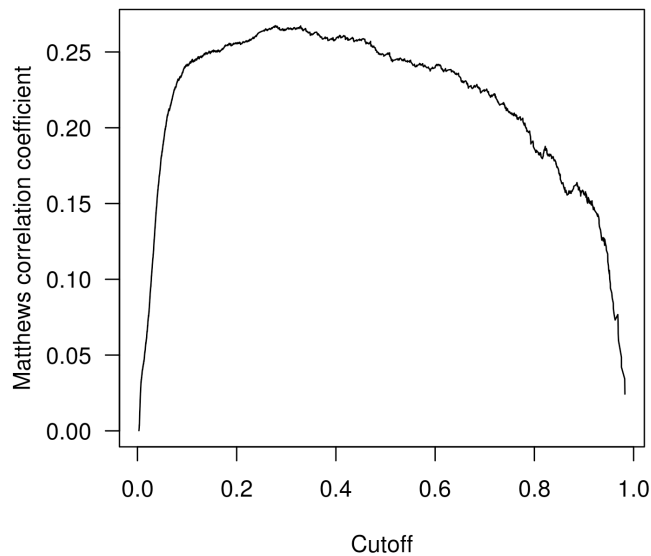

Fig.3. Matthews correlation coefficient for logistic regression for different values of probability threshold.

So-called magic features which are based on the ID of samples were considered by competition participants [5,6]. These features improved scoring essentially. We took the magic features presented in the competition forum post [7] and calculated ROC curve (Fig.5) and Matthews correlation coefficient (Fig.6) for different features sets. Features set 2 is the features set 1 with added magic features. For features set 2, we received AUC=0.91.

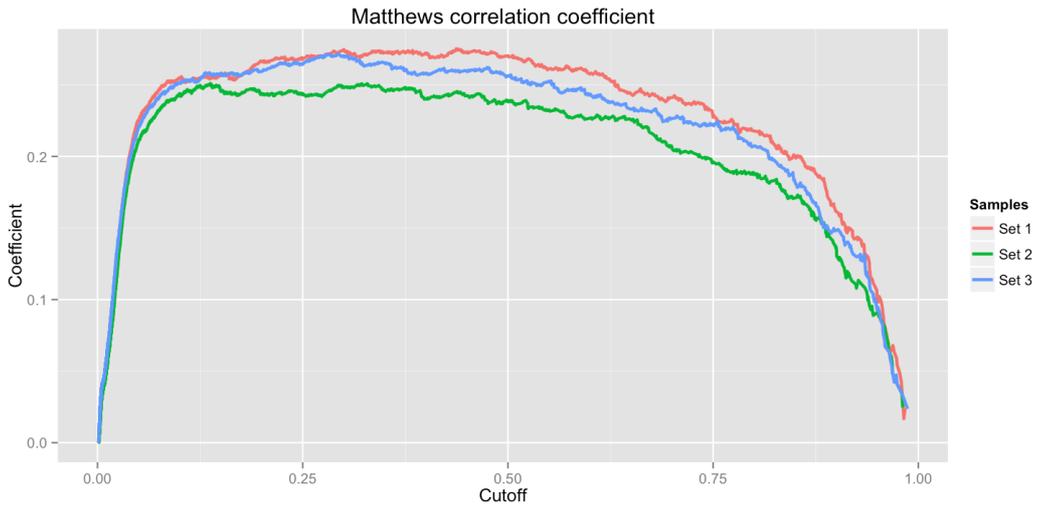

Fig.4. Matthews correlation coefficient for different samples sets.

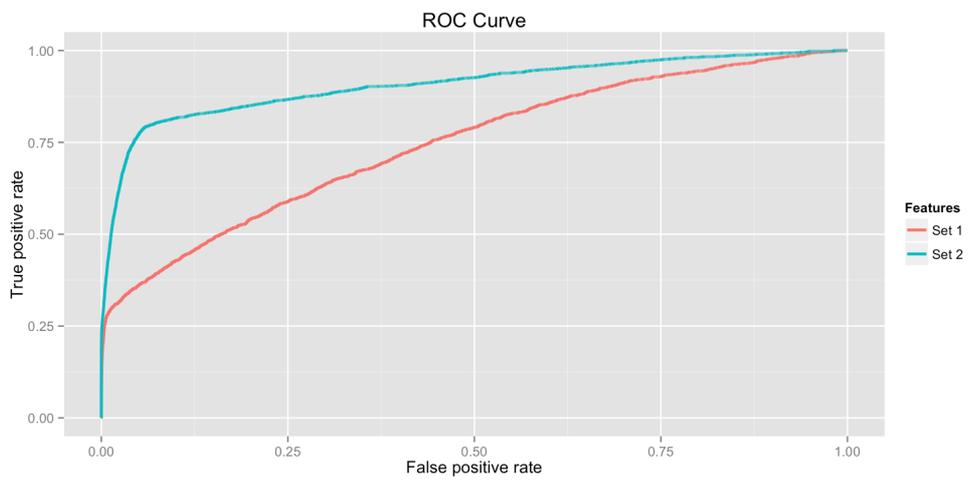

Fig.5. ROC curve for classification results for different sets of features (Set 2 is Set 1 with added magic features).

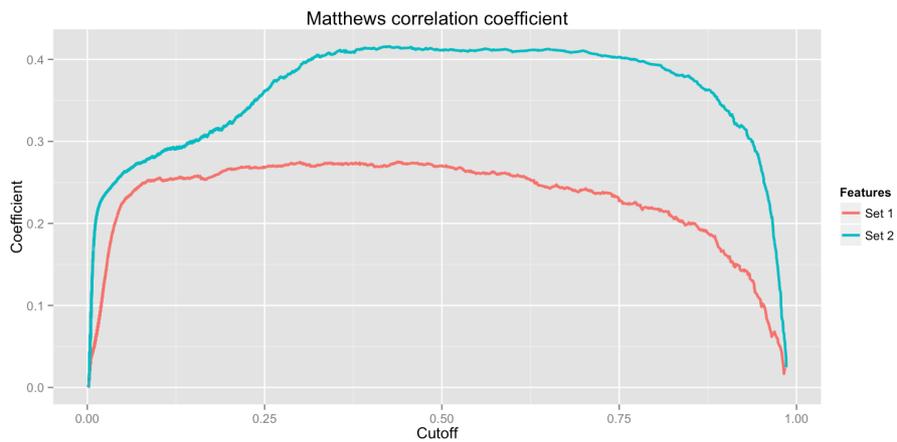

Fig.6. Matthews correlation coefficient for different sets of features.

## GENERALIZED LINEAR MODEL

Logistic regression can be considered as a special case of generalized linear model. Such type of logistic regression enables us to investigate the influence of numeric factors on the binary response. For this model, we can receive the coefficients for each feature under study. The data set under consideration has a large number of not available measure values (NA) for each sample. It is not a big problem for modern machine learning classifiers but when we try to apply the linear logistic regression, we need to preprocess the data to avoid NA values in the data set. The possible reason of such a large number of NA values is that in the data set, there are the manufacturing parts of different types. For one type of the parts, one set of measures is applied for quality controlling, for the other type of the parts, the other set of measures is applied. Taking into account that all measure features in the data set are represented as columns, the measures, which are not applied to some part, will have NA value. So, at first in the linear analysis, we need to group the parts by their types. Grouping can be performed by clusterization. Since the features in the data set are anonymized, we used only numeric features to study the possibility of applying the generalized linear logistic regression for this data set. We used the k-mean algorithm for the clusterization. We set up the value 0 for not available values of features in the data sets and value 1 for the features with numeric values of measure. To speed up the calculation, we chose 850 random features. Fig.7 shows the dependency of total within-clusters sum of squares from the number of clusters. The results show that the optimal number of groups of the parts is approximately 20-30.

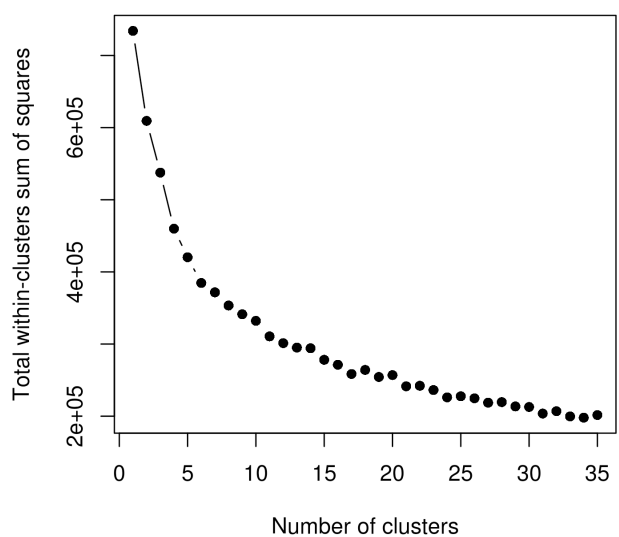

Fig.7. Total within-clusters sum of square from number of clusters.

We took the value 25 for the number of clusters. For the linear regression, we took the parts from one arbitrary cluster only. We removed the columns where there were more than 10% of NA values. We also removed the rows with more than 10% of NA values. The rest NA values were replaced by median value, calculated for appropriate numeric features. For the linear logistic regression, we used R package "glmnet" [8,9,10,11]. We used the logistic regression with LASSO regularization. The *glmnet* algorithms use cyclical coordinate descent, which optimizes successively the objective function over each parameter with the others fixed, and cycles repeatedly until the convergence [8]. Let us denote the probability that if the response is 1 by $p_1$, then the probability that the response is 0 will be $p_0=1-p_1$. For the logistic regression, we can write

$$\log\left(\frac{p_1}{p_0}\right) = \beta_0 + \beta^T x$$

$$p_1 = \frac{\exp(\beta_0 + \beta^T x)}{1+\exp(\beta_0 + \beta^T x)}$$

The coefficients $\beta, \beta^T$ can be found via minimizing the appropriate objective function. To find the optimal value for regularizing the *Lambda* parameter, we used the cross validation approach. Fig.8 shows the dependence of logarithmic value of *Lambda* from AUC value. The axis above indicates the number of nonzero coefficients for current *Lambda*, which is the effective degree of freedom for the LASSO model. Having this dependence, we can find the optimal value of regularization parameter, under which we can receive the best fitting.

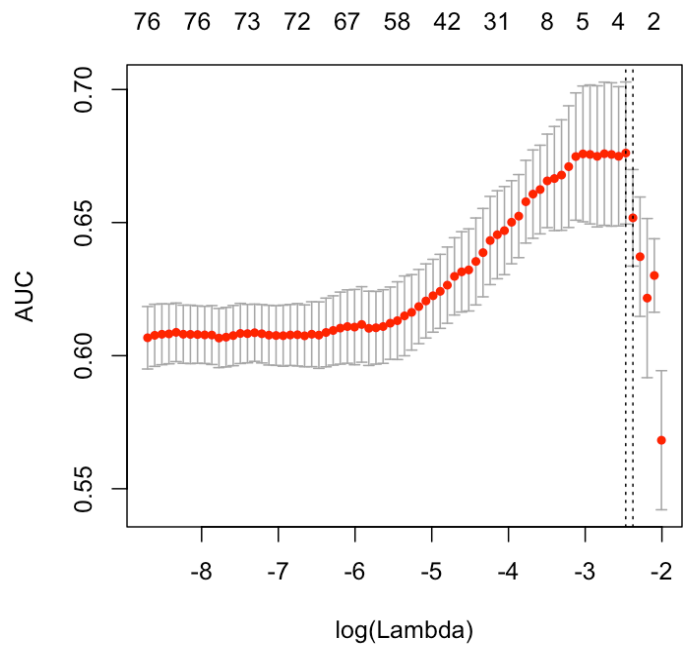

Fig.8. Dependence of Lambda from AUC value.

Using *Lambda=0.03*, we calculated nonzero coefficients of the generalized linear model for logistic regression which is shown on Fig.9.

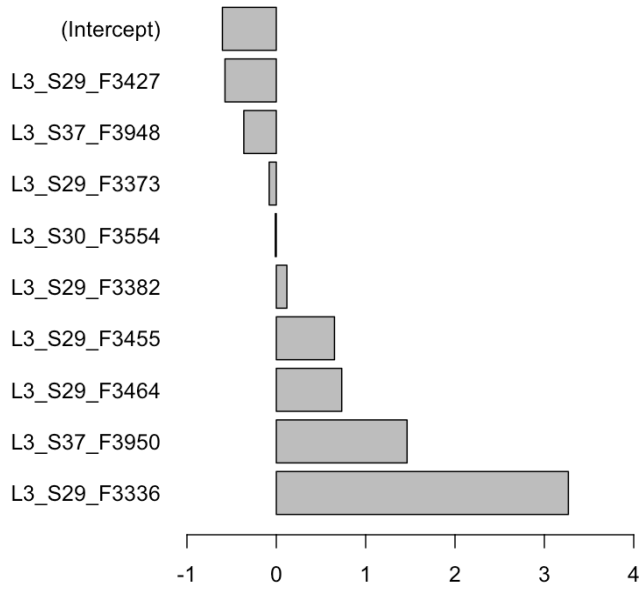

Fig.9. Coefficients of the generalized linear model for logistic regression.

Fig. 10 shows the histograms, correlation coefficients, pairs scatterplots for some chosen features. This figure shows the character of dependencies between these features.

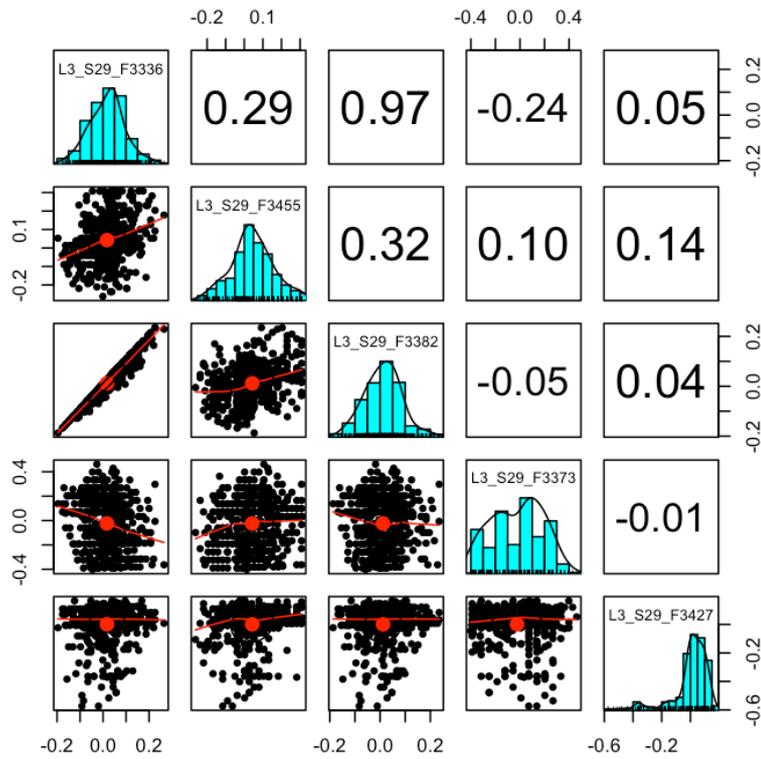

Fig.10. Histograms, correlation coefficients, pairs scatterplots for features.

I. BAYESIAN MODEL

For Bayesian model, we took the features which were found in the generalized linear model using LASSO regularization. The analysis was conducted using JAGS sampler software with "rjags" R package [12,13]. For modeling, we used the logistic regression. The idea is that a linear combination of metric predictors is mapped to the probability value via the logistic function, and the predicted 0 and 1 are Bernoulli distributed [12]. We can describe logistic regression as

$$p = Logistic(b_o + b_1 x_1 + b_2 x_2 + ... b_n x_n)$$

$$y \sim Bernulli(p)$$

$$\text{where } Logistic(x) = \frac{1}{(1+\exp(-x))}$$

The simple probabilistic model for logistic regression using BUGS syntax has the following code:

```
model{
  for (i in 1:n) {
    y[i] ~ dbern(p[i])
    logit(p[i]) <- b0+inprod(b[ ],x[i,])
  }
  b0 ~ dnorm(0,0.0001)
  for (j in 1:nfeat) {
    b[j] ~ dnorm(0,0.0001)
  }
}
```

Trace plots of samples vs the simulation index can be very useful in convergence assessment. Fig.11 shows the trace plot for b0 parameter of logistic model.

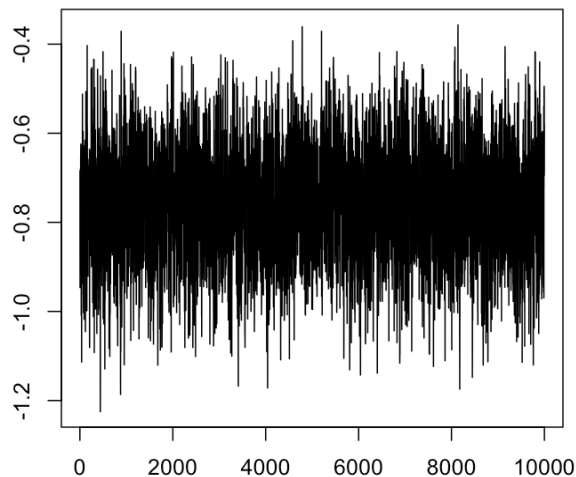

Fig. 11. Trace plot for Intercept parameter.

The trace plot demonstrates the stationary process, which means good convergence. Fig. 12 shows the probability density distributions for Intercept (b0) parameter.

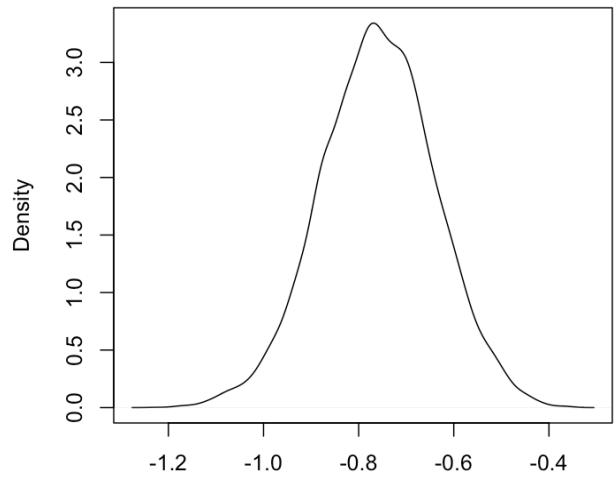

Fig.12. Density of distributions for Intercept parameter.

Fig.13 shows the examples for box plots for some logistic regression coefficients. The points on the figure denote the values of coefficients calculated using generalized linear model without regularization (*Lambda=0*) which was applied to the same features as in the case of Bayesian model.

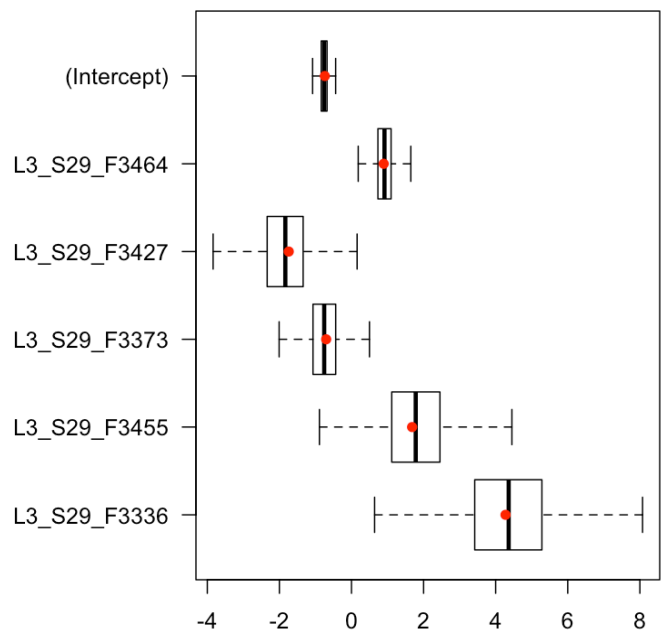

Fig.13. Box plots for logistic regression coefficients.

As the case study shows, the use of Bayesian approach allows us to model stochastic dependencies between different factors and receive the distributions for model parameters. Such an approach can be useful for assessing different risks related to quality control problems.

COMBINING MACHINE LEARNING WITH LINEAR AND BAYESIAN MODELS

One of the effective approaches of machine learning classification and regression is stacking, when predictions of models on one level are used as features for models on the next level. Using multilevel models with stacking approach is very popular among the participants of Kaggle competitions. The team where I was a participant used such an approach on the Kaggle competition Gruppo Bimbo Inventory Demand which we won. One can find our first place solution on that competition at [14]. Combining machine learning models and linear or Bayesian models on different levels can give us improved results for logistic regression. Fig. 14 shows a diagram of such possible multilevel model. On the first level, there are different XGBoost classifiers with different sets of features and subsets of samples, on the second level, probabilities predicted on the first level can be blended with appropriate weights using linear or Bayesian regression.

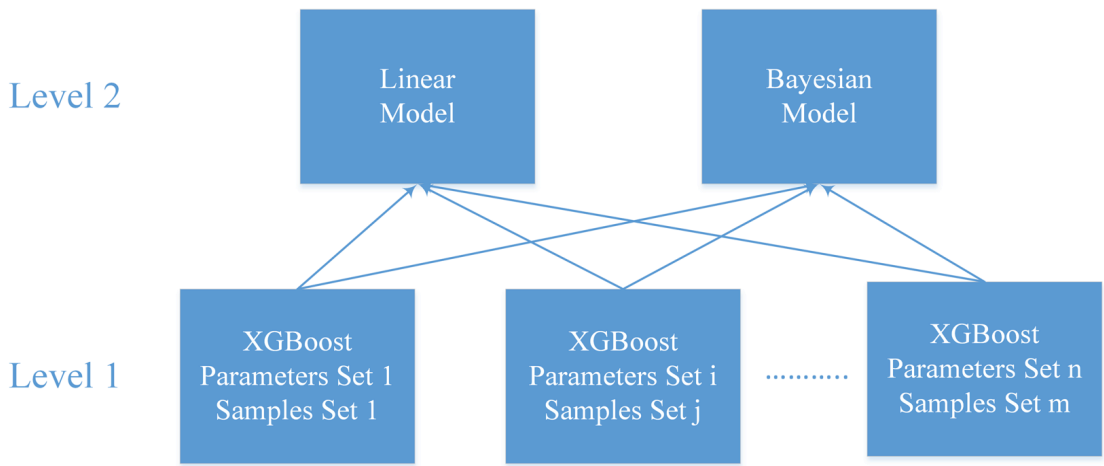

Fig.14. Logistic regression with machine learning models on the first level and linear and Bayesian models on the second level.

Let us consider the use of generalized linear model for logistic regression with independent variables which are the probabilities predicted by XGBoost models on the first levels. We used different sets of parameters for 3 XGBoost models, they are - set 1: max.depth = 15, colsample_bytree = 0.7; set 2: max.depth = 5, colsample_bytree = 0.7; set 3: max.depth = 15, colsample_bytree = 0.3. For these 3 models, we used the same subset of samples. Fig.15, 16 show the dependence of Matthews correlation coefficient from a probability threshold for different subsets of features, where features set 2 is features set 1 with 4 added magic features, mentioned above. For Bayesian models, we used the same 3 subsets of parameters with different subsets of samples. As it was shown above, for different samples subsets, we received slightly different results for Matthews correlation coefficient (Fig.6). These differences can be taken into account while using Bayesian model. We used Bayesian model for logistic regression, as covariates we used the probabilities predicted for three XGBoost models and for 3 different subsets of samples. Fig.17 shows the density of distributions for coefficients of probabilities predicted by different XGBoost models. Fig.18 shows boxplots for these coefficients.

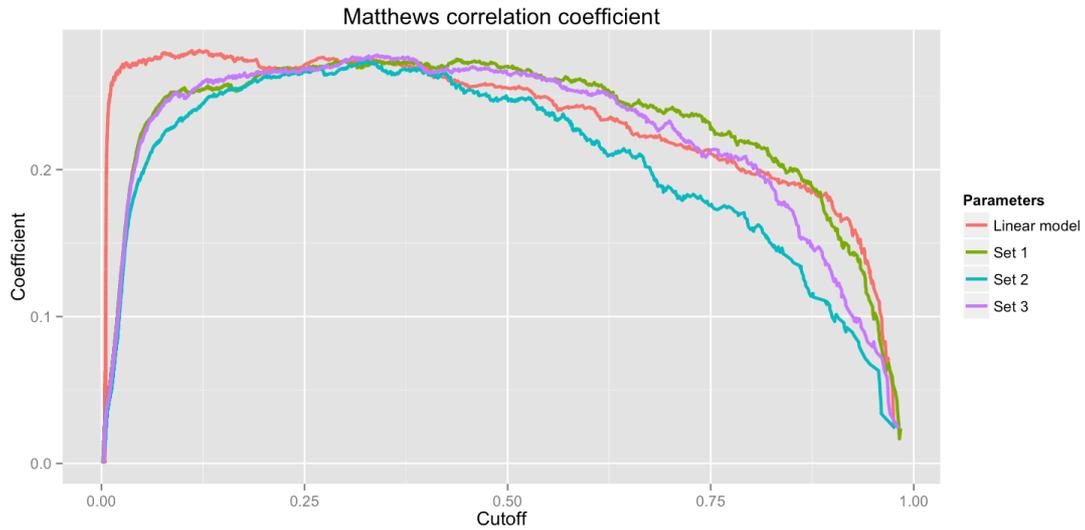

Fig.15. Matthews correlation coefficient for different XGBoost parameter sets (features set 1)

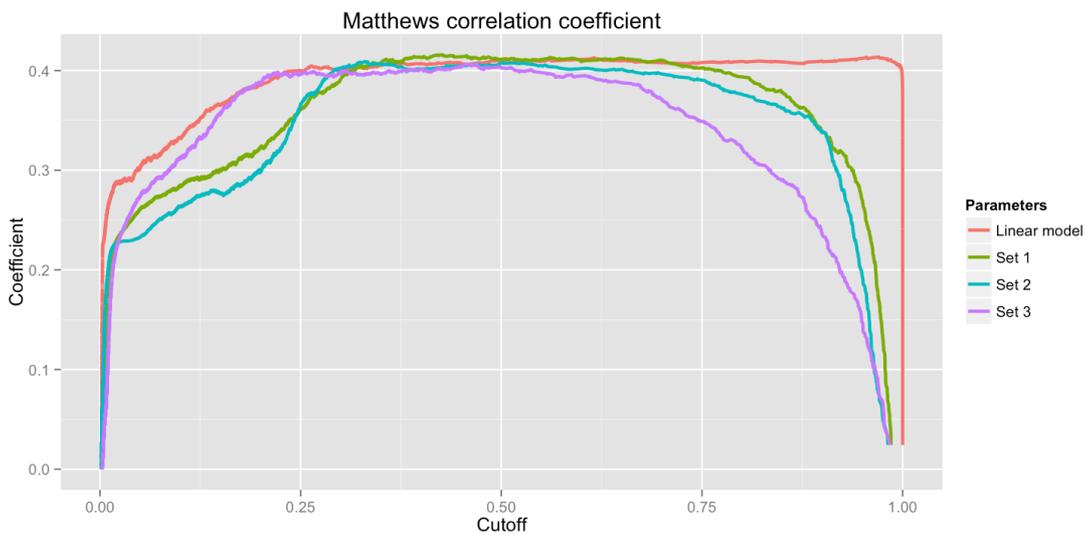

Fig.16. Matthews correlation coefficient for different XGBoost parameter sets (features set 2)

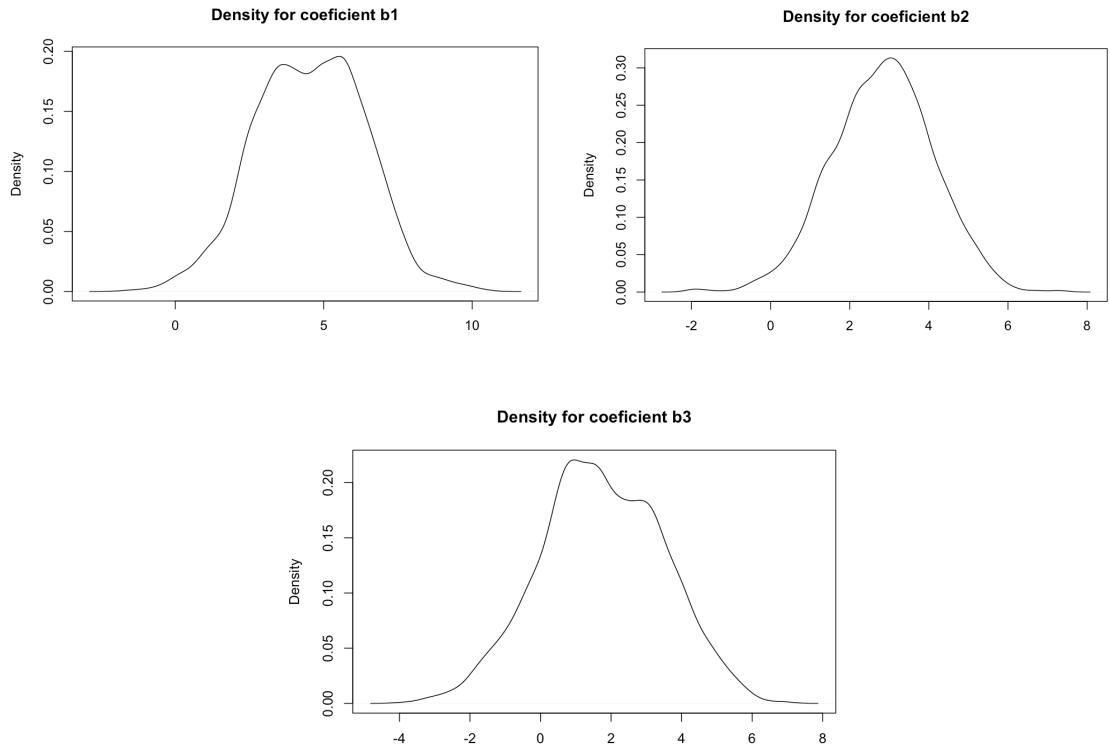

Fig.17. Density of distributions for coefficients of probabilities predicted by different XGBoost models.

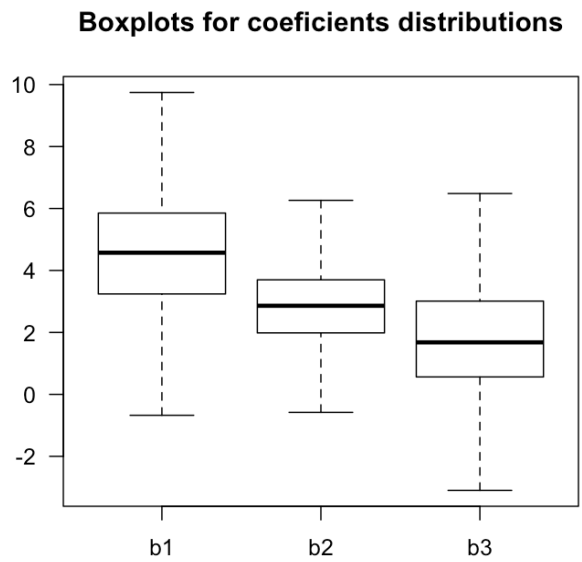

Fig.18. Boxplots for coefficients of probabilities predicted by different XGBoost models.

# STUDY OF RELIABILITY OF PARTS

One of the possible usages of measurements from manufacturing assembly line is the prediction of working lifetime for parts and devices. We did not have such historical data for the lifetime of working parts, so we tried to simulate such an approach. The working lifetime of parts can be described by Weibull distribution

$$f(t) = \frac{\beta}{\alpha}\left(\frac{t}{a}\right)^{\beta-1} \exp\left[-\left(\frac{t}{\alpha}\right)^{\beta}\right]$$

Let us suppose that we have 3 measurements, which can be used for predicting the lifetime of some part. Suppose the scale parameter $\alpha$ of Weibull distribution has the linear dependence from these measurements:

$$\alpha = b_0 + b_1 x_1 + b_2 x_2 + b_3 x_3$$

The parameters of this model can be found using the data e.g. from the repair centers. Having id# of broken parts, their lifetime and their measurements on the manufacturing line, one can study the possibility to predict the distribution of lifetime for these parts using measurements from the manufacturing line as predictors. We simulated the Weibull distribution with arbitrary chosen parameters of the model. Fig. 18 shows such simulated Weibull distribution. Then the received data were used as input data for Bayesian logistic regression. Fig.20 shows the density for the $\beta$ parameter of Weibull distribution and boxplots for the coefficients of Bayessian logistic regression model. Having these distributions, one can find the distribution for the lifetime of parts under study. Of course, these are only the simulation results. Having real data, this approach can be effective, combining the machine learning model on the first level and probabilistic Bayesian model on the second level.

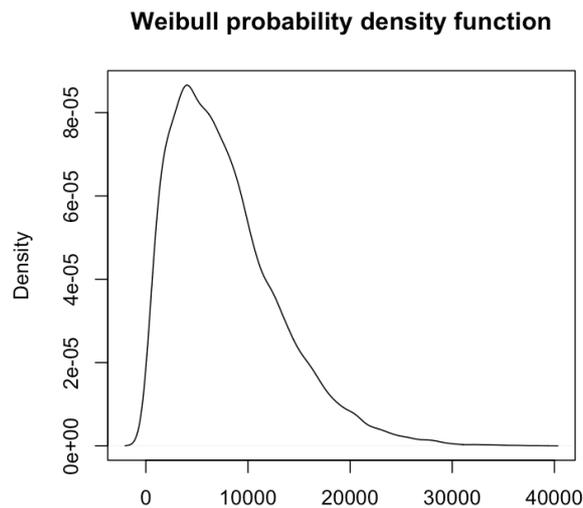

Fig.19. Simulated Weibull distribution

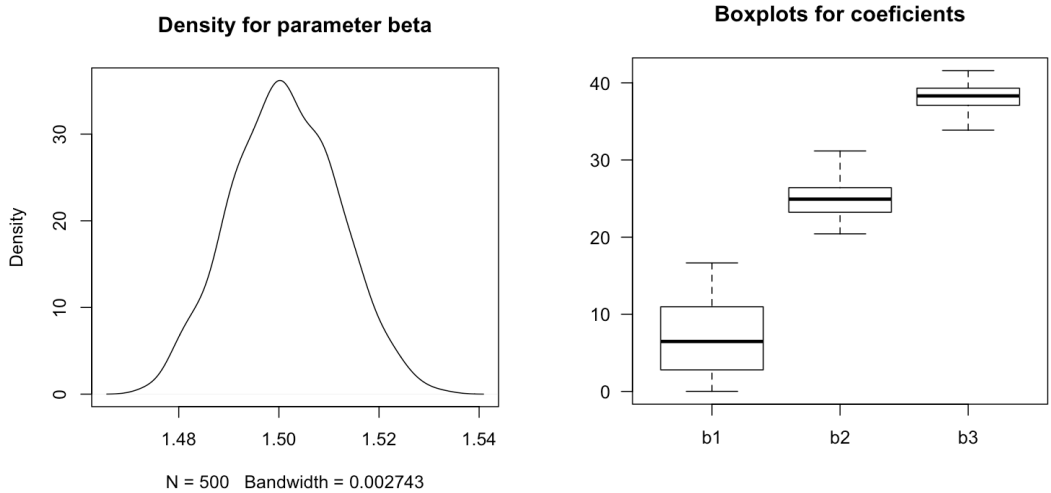

Fig.20. Density for $\beta$ parameter of Weibull distribution and boxplots for coefficients of Bayesian logistic regression model.

Conclusion

In our case study, we showed different approaches for the logistic regression in the issue of manufacturing failures detection. Machine learning approach can give us the best-scored failure detection. The generalized linear model for the logistic regression makes it possible to investigate influence factors on the failure detection in the groups of manufacturing parts. Using Bayesian model, it is possible to receive the statistical distribution of model parameters, which can be used in the risk assessment analysis. Using 2-level models, we can receive more precise results. Using Bayesian model on the second level with the covariates that are the probabilities predicted by machine learning models on the first level, makes it possible to take into account the differences in results for machine learning models received for different sets of parameters and subsets of samples in case of highly imbalanced classes.

Acknowledgement

Special thanks to Bosch company for organizing so interesting Kaggle competition "Bosch Production Line Performance" and for awarding me the travel grant for attending the IEEE BigData 2016 conference !

[1] Kaggle competition "Bosch Production Line Performance". URL: https://www.kaggle.com/c/bosch-production-line-performance
[2] Chen, Tianqi, and Carlos Guestrin. "XGBoost: A Scalable Tree Boosting System." *arXiv preprint arXiv:1603.02754* (2016).
[3] J. Friedman. "Greedy function approximation: a gradient boosting machine.", *Annals of Statistics,* 29(5):1189–1232, 2001.
[4] J. Friedman. "Stochastic gradient boosting.", *Computational Statistics & Data Analysis*, 38(4):367–378, 2002.

[5] Kaggle competition "Bosch Production Line Performance". The Magical Feature : from LB 0.3- to 0.4+. URL:https://www.kaggle.com/c/bosch-production-line-performance/forums/t/24065/the-magical-feature-from-lb-0-3-to-0-4

[6] Kaggle competition "Bosch Production Line Performance". Road-2-0.4+. URL:https://www.kaggle.com/mmueller/bosch-production-line-performance/road-2-0-4

[7] Kaggle competition "Bosch Production Line Performance". Road-2-0.4+ --> FeatureSet++ . URL: https://www.kaggle.com/alexxanderlarko/bosch-production-line-performance/road-2-0-4-featureset

[8] Jerome Friedman, Trevor Hastie and Rob Tibshirani. (2008). Regularization Paths for Generalized Linear Models via Coordinate Descent. *Journal of Statistical Software*, Vol. 33(1), 1-22 Feb 2010.

[9] Noah Simon, Jerome Friedman, Trevor Hastie and Rob Tibshirani. (2011). Regularization Paths for Cox's Proportional Hazards Model via Coordinate Descent. *Journal of Statistical Software*, Vol. 39(5) 1-13.

[10] Robert Tibshirani, Jacob Bien, Jerome Friedman, Trevor Hastie, Noah Simon, Jonathan Taylor, Ryan J. Tibshirani. (2010). Strong Rules for Discarding Predictors in Lasso-type Problems. *Journal of the Royal Statistical Society: Series B (Statistical Methodology)*, 74(2), 245-266.

[11] *Stanford Statistics Technical ReportGlmnet Vignette*. URL: http://www.stanford.edu/~hastie/glmnet/glmnet_alpha.html

[12] Kruschke, John. *Doing Bayesian data analysis: A tutorial with R, JAGS, and Stan*. Academic Press, 2014.

[13] Martyn Plummer. JAGS Version 3.4.0 user manual. URL:http://sourceforge.net/projects/mcmcjags/files/Manuals/3.x/jags_user_manual.pdf

[14] Kaggle competition "Grupo Bimbo Inventory Demand" #1 Place Solution of The Slippery Appraisals team. URL:http://www.kaggle.com/c/grupo-bimbo-inventory-demand/forums/t/23863/1-place-solution-of-the-slippery-appraisals-team